\documentclass{article}
\usepackage{ijcai21}

\usepackage[utf8]{inputenc} 
\usepackage[T1]{fontenc}    
\usepackage[hidelinks]{hyperref}       
\usepackage{url}            
\usepackage{booktabs}       
\usepackage{amsfonts}       
\usepackage{nicefrac}       
\usepackage{microtype}      

\usepackage{epsfig}
\usepackage{amssymb}
\usepackage{soul}
\usepackage{times}  
\usepackage{helvet} 
\usepackage{courier}  
\usepackage{graphicx}  
\usepackage{caption}
\usepackage{subcaption}
\usepackage{amsmath}
\usepackage{amsthm}
\usepackage{cases}
\usepackage{color}
\usepackage{xspace}
\usepackage[ruled,vlined]{algorithm2e}   
\usepackage[noend]{algpseudocode}
\algnewcommand\And{\textbf{and}}
\usepackage[dvipsnames,table,xcdraw]{xcolor}

\usepackage{multirow}

\newif\ifdebugdoc\debugdoctrue
\ifdebugdoc

\newcommand{\todo}[1]{\textbf{\textcolor{red}{[TODO: #1]}}}
\newcommand{\remind}[1]{\textit{ \textcolor{red}{[Remind: #1]}}}
\newcommand{\note}[1]{\vskip 2ex \noindent\colorbox{yellow}{\parbox{\columnwidth}{#1}}\vskip 2ex}
\newcommand{\q}[1]{\vskip 2ex \noindent\colorbox{cyan}{\parbox{\columnwidth}{\textbf{Question:} #1}}\vskip 2ex}
\newcommand{\idea}[1]{\vskip 2ex \noindent\colorbox{magenta}{\parbox{\columnwidth}{\textbf{Idea:} #1}}\vskip 2ex}
\newcommand{\ys}[1]{\colorbox{blue}{Yuanchao:} \textcolor{blue}{#1}}
\newcommand{\fwho}[2]{\colorbox{red}{#2:} #1}

\newcommand{\del}[1]{\textcolor{blue}{\sout{#1}}}

\else
\newcommand{\todo}[1]{}
\newcommand{\remind}[1]{}
\newcommand{\note}[1]{}
\newcommand{\q}[1]{}
\newcommand{\idea}[1]{}
\newcommand{\ys}[1]{}
\newcommand{\fwho}[2]{}

\newcommand{\del}[1]{}

\fi

\title{Custom Object Detection via Multi-Camera Self-Supervised Learning}
\author{
	Yan Lu$^1$\footnote{Contact Author}\And
	Yuanchao Shu$^2$
	\affiliations
	$^1$New York University\\
	$^2$Microsoft Research\\
	\emails
	jasonengineer@hotmail.com,
	yuanchao.shu@microsoft.com
}

\newcommand{\name}[0]{{\sc MCSSL}\xspace}
\newcommand{\aka}{{\it a.k.a.,}\xspace}
\newcommand{\eg}{{\it e.g.,}\xspace}
\newcommand{\ie}{{\it i.e.,}\xspace}

\newcommand{\bbox}{{bounding box}\xspace}
\newcommand{\bboxes}{{bounding boxes}\xspace}
\newcommand{\cs}{{camera-specific}\xspace}
\newcommand{\ncs}{{non-camera-specific}\xspace}

\begin{document}

\maketitle

\begin{abstract}
	This paper proposes \name, a self-supervised learning approach for building custom object detection models in multi-camera networks. \name associates \bboxes between cameras with overlapping fields of view by leveraging epipolar geometry and state-of-the-art tracking and reID algorithms, and prudently generates two sets of pseudo-labels to fine-tune backbone and detection networks respectively in an object detection model. To train effectively on pseudo-labels, a powerful reID-like pretext task with consistency loss is constructed for model customization. Our evaluation shows that compared with legacy self-training methods, \name\ improves average mAP by $5.44\%$ and $6.76\%$ on WildTrack and CityFlow dataset, respectively.
\end{abstract}

\section{Introduction}

Object detection plays a pivotal role in video analytics. Although deep neural network (DNN)-based object detection models pre-trained on large public datasets (\eg MS-COCO) exhibit decent performance in various scenarios, custom models are more desired due to its higher accuracy and robustness~\cite{2016backbone1,2019finetune1}. 

Model customization relies on context-specific (or domain-specific) training data. Unlike general-purpose training datasets, large-scale context-specific labels are way harder to collect in a sustainable manner. For most computer vision tasks, manual labeling (\eg draw bounding boxes) remains to be the major source of training data. Nonetheless, human annotation is known to be costly and time-consuming, and distributing frames outside of the camera network also raises privacy concerns. While we are witnessing advancements in semi-supervised and weakly-supervised learning, performance of models trained from the state-of-the art semi- and weakly-supervised algorithms still falls short of supervised object detectors. In semi-supervised learning, for instance, it's hard to converge on regularization terms (\eg consistency loss)~\cite{2019consistencylimit} in large-scale unlabeled data. Similarly, detection performance suffers from local minimums in weakly-supervised approaches~\cite{2018domainadaption2}.


This paper is motivated by a simple question, \emph{can we improve self-supervised object detection with a fleet of cameras that have (partially) shared field-of-view (FOV)?} Driven by advances in computer vision and the plummeting costs of
camera hardware, organizations are deploying video cameras at scale for a variety of applications. In many scenarios, cameras have overlapping areas for the spatial monitoring of physical premises. Detection results on neighboring cameras could potentially serve as pseudo-labels and allow each camera to learn its own detector continuously.

To achieve this goal, however, we need to tackle the following two major challenges.

\textbf{How to create pseudo-labels?} Object detection accuracy varies for cameras and changes over time due to traffic dynamics and environmental factors (\eg lighting changes), making it hard to identify high-quality pseudo-labels. It is equally important to assign the right set of pseudo-labels to camera(s) as blindly training with all pseudo-labels adversely impacts model customization. 

\textbf{How to learn from noisy labeled data?} A straightforward way to customize model is to fine-tune the default object detection model on each camera using pseudo-labels. However, fine-tuning a large DNN with insufficient and noisy pseudo-labels tends to lead to accuracy drop and overfitting~\cite{hendrycks2018glc}. 

To address these issues, we propose \name, a novel self-supervised training mechanism to customize object detection models on each camera. The key idea of \name is to create pseudo-labels at different confidence levels and use them to train different parts of the network separately. \name divides object detection models into two parts: \emph{backbone network} which provides discriminative low-level feature maps, and \emph{detection network} which consists of region proposal network (RPN), RoI feature extractor etc. The two parts have their unique characteristics (Table~\ref{table:designchoice}). Backbone network has way more parameters than RPN and RoI, hence it demands more training data to fine-tune itself. On the other hand, it is less susceptible to training data noises since it is trained for low-level features~\cite{2020backbone1,2018backbone2,2016backbone1}. We also found that, for most off-the-shelf object detection models, \bboxes with high classification score are rarely false positives. However, it is common to see false negative cases with low classification score. 
\begin{table}[!htb]
	\begin{center}
		{\footnotesize
			\begin{tabular}{|c|c|c|c|c|c|c|c|c|}
				\hline
				\hline
				& Training Data Demand & Noise Sensitivity \\
				\hline
				Detection Network & Low & High \\
				\hline
				Backbone Network & High & Low \\
				\hline
				\hline
		\end{tabular}}
	\end{center}
	\caption{Object Detection Model Characteristics.}\label{table:designchoice}
	\vspace{-10pt}
\end{table}

Based on these two insights, \name categorizes \bboxes detected by the base model on each camera as \emph{confident pseudo-labels} and \emph{uncertain pseudo-labels} based on their classification score. Despite relatively low volume, \emph{confident pseudo-labels} serve nicely for detection network fine-tuning due to their high quality. On the contrary, a larger number of \emph{uncertain pseudo-labels} fit well with the backbone network which is more tolerable to noises. 

To build custom object detection model, detection network and backbone network on each camera need to be trained independently with its own context-specific data. To this end, \name adopts advanced video reID algorithms to associate \bboxes across cameras. To improve the efficiency and precision of \bbox matching, a prune-and-augment approach is introduced by leveraging epipolar constraints and tracking. Akin to feature categorization in classification and reID tasks~\cite{2018reid1,2018reid3}, we treat paired \bboxes as \emph{\ncs data} due to their camera-invariant features on the same object. Combined with the above finding, \emph{uncertain pseudo-labels} which belong to \emph{\ncs data} is used to update the backbone network on each camera. To work with pairs of images as input, we devise a novel reID-like pretext task, turning backbone network fine-tuning to consistency training. When we get a more powerful backbone network, we use traditional self-training framework to update detection network (\ie RPN and RoI feature extractor) with the entire set of \emph{confident pseudo-labels}.

We found the results are promising. Compared with the best self-training algorithm, \name, on average, improves mAP by $5.44\%$ and $6.76\%$ for each camera on WildTrack and CityFlow dataset, respectively. Two out of seven cameras even achieve an mAP gain of as high as $10\%$ on the WildTrack dataset. In summary, we made the following three intellectual contributions. 

i) We proposed the first self-supervised learning approach that allows object detection model customization in multi-camera networks. 

ii) We devised an effective approach to generate training data and fine-tune different parts of an object detection model with a reID-like pretext task. 

iii) We achieved the new state-of-the-art mAP results of self-supervised learning-based custom object detection on WildTrack and CityFlow datasets.

\section{Related Works}

\textbf{Anchor-based object detection}: 
Anchor-based deep object detection models \cite{yolov3,2017maskrcnn,2015fasterrcnn,2019m2det} comprise of three modules: 1) Backbone network which extracts general features (\ie edges, corners) of an given image; 2) Region proposal network (RPN) \cite{2015fasterrcnn} that generates candidate bounding boxes based on simpler components from lower layers in backbone network; and 3) RoI feature extractor, which extracts fine-grained features and assigns class probability for each RoI generated by RPN. Models contain all three modules are called three-stage object detection models (\eg Mask RCNN) whereas two-stage models (\eg YOLOv3 and M2Det)\cite{yolov3,2019m2det} remove RPN and run RoI feature extractor directly on feature blocks generated by backbone network to improve inference speed. 
\name builds on top the existing anchor-based object detection architecture and fine-tunes backbone network and detection network independently with prudently generated training data for model customization.

\textbf{Semi- and weakly-supervised object detection}: Despites recent advancements~\cite{2019RCNN,2019selfOD1,2018domainadaption1,2018domainadaption2,2019consistency1}, today's semi- and weakly-supervised learning algorithms still fall short on accuracy in object detection tasks. 
A common approach to semi-supervised object detection is mining-training. \cite{2019RCNN} is the first end-to-end semi-supervised framework for object detection. To utilize a large amount of unlabeled data and handle label noises, many works~\cite{2019selfOD1,2019selfOD2,2019selfOD3} seek to construct auxiliary tasks (\aka pretext tasks) to indirectly train the network. For example, \cite{2019selfOD2,2019selfOD3} add a knowledge graph mining task, and \cite{2019selfOD1} constructs three new labeling tasks (closeness labeling, multi-object labeling and foreground labeling) to assist object detection model training. These kinds of approaches are also known as self-supervised learning as in auxiliary tasks, pseudo-labels are found or mined in unlabeled data automatically. Besides self-supervised approaches, constructing consistency \cite{2019consistency1} between different versions of a given image became an effective tool for enhancing detection models' performance on unlabeled data. 
Inspired by self-supervised learning and consistency learning approaches, we devise a new reID-like pretext task trained by means of consistency learning to assist custom object detection model training on multi-camera datasets.


\textbf{Multi-camera detection}: 
To deal with occluded objects from a single view, many works~\cite{2017detect,2017multiview} utilize multi-view streams to build powerful 3D detection algorithms for all cameras. 
However, the majority of the methods are supervised learning-based, requiring a significantly more labeled data than monocular object detection. More recently, we have seen works on multi-view human pose estimation~\cite{2019view1} that do not learn a 3D model but instead seek to train 2D models for each camera through adding a new self-supervised learning task~\cite{2019view1}. 
To the best of our knowledge, \name is the first self-supervised learning-based approach to get custom object detection models in multi-camera environments.


\section{Design}
\label{sec:design}

\begin{figure*}[!htb]
	\centering
	\includegraphics[width=2.0\columnwidth]{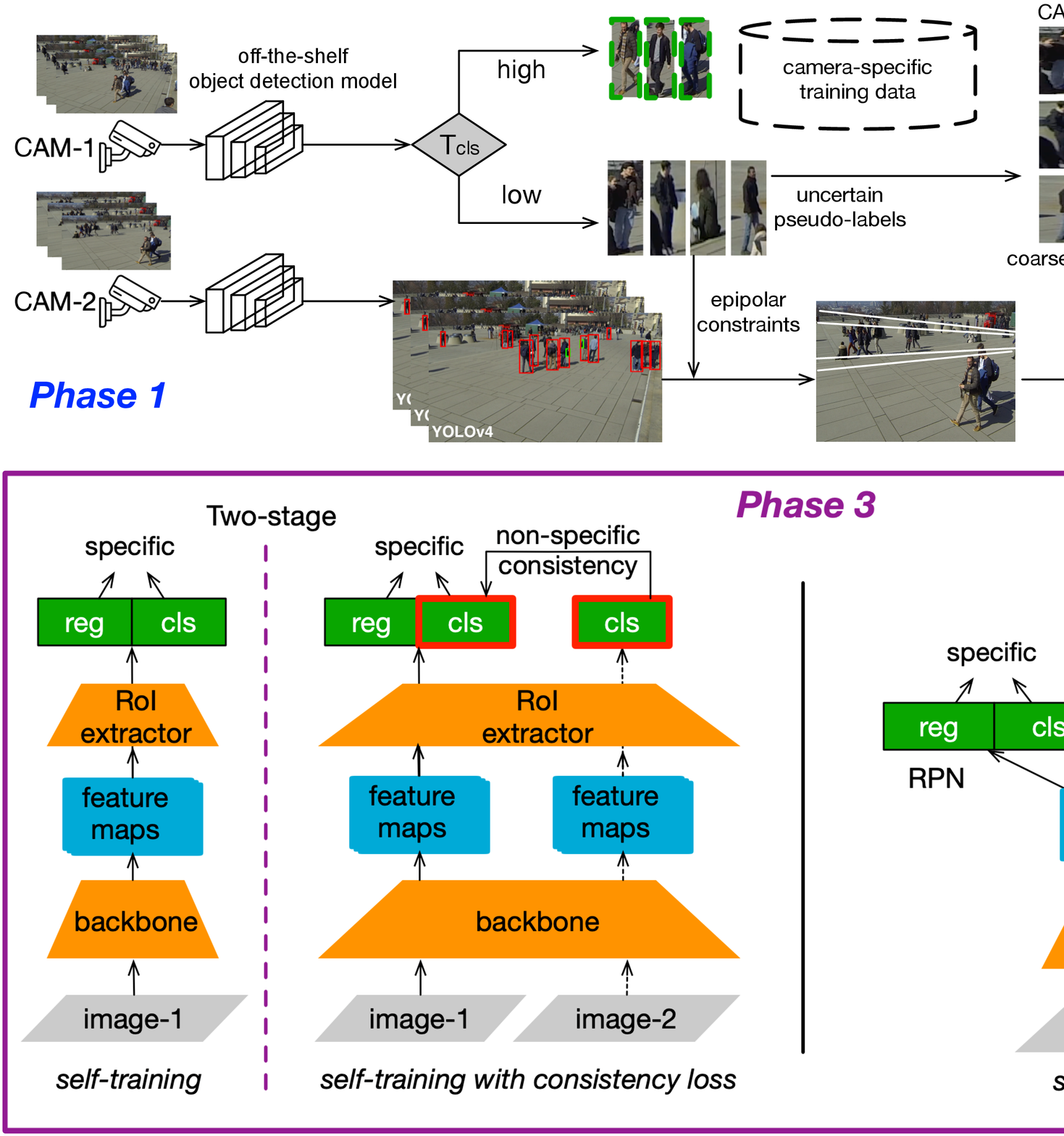} 
	\caption{\name overview.}
	\vspace{-10pt}
	\label{fig:new_design}
\end{figure*}

\name\ works in multi-camera scenarios where at least two cameras (partially) share field-of-view. At beginning, cameras are running off-the-shelf object detection models (\ie base model) trained on large-scale public dataset (\eg YOLOv3, Mask R-CNN). The goal of \name\ is to build accurate custom object detection model for each camera over time through cross-camera model fine-tuning. In what follows, we use two cameras as a simple example to elaborate the process of model customization on $cam^1$ with the help of $cam^2$ (Figure~\ref{fig:new_design}).

\textbf{Phase 1: pseudo-label generation}: At first, object detection results are obtained from the base model on frames from both $cam^1$ and $cam^2$. \name\ sets \bboxes with high classification score (\eg $0.8$) as \emph{confident pseudo-labels} and the remaining \bboxes as \emph{uncertain pseudo-labels}. 

\textbf{Phase 2: cross-camera pseudo-label sharing}: \name treats views of one object on different cameras as styles-transfered images, and associates pseudo-labels across cameras using state-of-the-art reID models. Notably, epipolar geometry and tracking are leveraged which significantly reduces compute overhead and improves accuracy. Associated \emph{pseudo-labels} are categorized as \emph{\ncs} and \emph{\cs} training data, where \emph{\ncs} training data refer to objects seen by multiple cameras whereas \emph{\cs} training data are objects appear on a single camera. 

\textbf{Phase 3: consistency learning}: \name constructs a reID-like pretext task that uses \ncs training data to fine-tune backbone network with consistency loss. Camera-specific training data is used to customize the detection network as in most existing self-training algorithms.


\subsection{Pseudo-label Generation}
\label{subsec:coarse-pseudo}
When $cam^1$ and $cam^2$ collect enough new images, we first use off-the-shelf object detection models to generate pseudo labels. We denote $BB^1$ and $BB^2$ pseudo-labels for $cam^1$ and $cam^2$, respectively. 
Intuitively, \bboxes with higher classification score are more likely to be true positive. As reported in a large body of work in computer vision~\cite{yolov3,2017maskrcnn,2015fasterrcnn}, using a high classification score to filter \bboxes tends to lead to high precision but low recall across the majority of DNN object detection models. That is, \bboxes with high classification score are rarely false positives whereas false negative \bboxes due to low classification score are commonly seen. 
Based on this insight, we set a high threshold $T_{cls}$, and use high-quality confident pseudo-labels $BB_c$, \bboxes whose classification scores are larger than $T_{cls}$, to train upper layer detection networks (\ie RoI feature extractor). Accordingly, uncertain pseudo-labels ($BB_u$) are used to train the initial CNN layers (\ie backbone network) which are less susceptible to noises~\cite{2016sensitivity,co2019procedural}.

\subsection{Cross-camera Pseudo-label Sharing}
\label{subsec:invariant-data}
Style transfer is a commonly used data augmentation technique in DNN model training~\cite{2018domainadaption1,2018domainadaption2}. A key advantage of multi-camera network is the richness of data from different vantage points. 
Hence, in \name, we associate \bboxes of the same object on different cameras and feed them in model fine-tuning. 

In a nutshell, \bbox association is achieved by reID. However, na\"ive reID poses two challenges. First, state-of-the-art reID models are only able to achieve mAP of around $65.3\%$, which leads to a decent number of false positives and impairs model fine-tuning. Second, pairwise comparison between \bboxes on all cameras incurs a non-linear computation overhead, which is prohibitively high for scenarios with busy traffic. To deal with these two issues, we employ a prune-and-augment approach, which first filters out a large number of \bboxes that are less likely to be confirmed by reID using epipolar constraints, and then augments refined pseudo-label pairs through tracking. 

\subsubsection{Epipolar Constraint-based Pruning}
\label{subsubsec:epipolar-geometry}
When two cameras view the same 3D space from different viewpoints, geometric relations among 3D points and their projections onto the 2D plane lead to constraints on the image points. This intrinsic projective geometry is captured by a fundamental matrix $F$ in epipolar geometry, which can be calculated as $F=K^{-T}_{2}[t]_{\times}RK^{-1}_{1}$ where $K_{1}$ and $K_{2}$ represent intrinsic parameters, and R and $[t]_{\times}$ are the relative camera rotation and translation which describe the location of the second camera relative to the first in global coordinates (\aka extrinsic parameters). Given $F$, for a physical 3D position $P$ in the overlapping area of $cam^{1}$ and $cam^{2}$, we have $p_{1}^{T}Fp_{2} = 0$, where $p_{1}$ and $p_{2}$ are the projected scene point from $P$ on $cam^{1}$ and $cam^{2}$. In essence, this equation characterizes an epipolar plane containing $P$ and epipoles $O_1$ and $O_2$ of both cameras.

\begin{figure}[h]
	\centering
	\includegraphics[width=0.5\columnwidth]{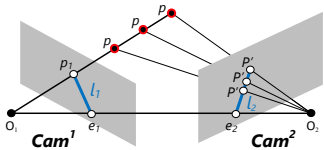} 
	\caption{Illustration of epipolar constraints.}
	\label{fig:epipolar_constraints}
\end{figure}

Epipolar plane offers a unique characteristic in building associations between \bboxes on different cameras. As can be seen from Figure~\ref{fig:epipolar_constraints}, the intersection of the epipolar plane with the image plane are two lines, which are called epipolar lines. This means for any particular point $p_1$ on $cam^{1}$, it is always mapped to a point alone the epipolar line $l_2$ in the image from $cam^{2}$. 

Given the epipolar constraints, we can now map a \bbox in the image from the ``teacher camera'' to four epipolar lines in another camera's image, which significantly reduces the search space of potential bounding boxes. For instance, it reduces the search space by 12x on the WildTrack dataset. Note that although explanations above assume cameras are calibrated and time-synchronized, we add a fudge factor in our spatial filtering algorithm to compensate calibration noises and slight time shift. 
Since epipolar geometry only defines an area for each \bbox, we set all \bboxes in $BB_u$ fall into the area as candidate \bboxes (\ie coarse \ncs training data). In order to fine-tune the object detection model of camera $i$, \name applies mapping on all cameras share the view with $i$. In the example of customizing an object detection model for $cam^1$ with the help of $cam^2$ (Figure~\ref{fig:new_design}), we run epipolar geometry-based mapping on all \bboxes in $cam^2$ to find all candidate \bboxes on $cam^1$.

\subsubsection{Data Augmentation with Tracking}
Epipolar constraints effectively reduces the search space of \bboxes for reID. Nonetheless, it filters out pairs of \bboxes on cameras at different times. For instance, $person_{A}$ on $cam^1$ at $i_{th}$ frame may not fall within the epipolar constraint of the \bbox of $person_{A}$ on $cam^2$ at $j_{th}$ frame, despite that this is a valid pair of non-camera-specific data. To revive this large set of training data (due to its combinatorial nature), we leverage temporal correlations on each camera to find \bboxes belong to the same object. In specific, once we find $person_{A}$ on $cam^1$ and $cam^2$ at the $i_{th}$ frame, we run SiamMask-E~\cite{chen2019fast}, a state-of-the-art tracking algorithm on subsequent four frames from both cameras to get \bboxes of $person_{A}$. This set of data is called ``coarse reID training data'' (Figure~\ref{fig:new_design}). 


Data augmentation with tracking also allows us to use video reID algorithm to finalize \bbox association. Compared with image reID, video reID~\cite{2019videoreid1,2019videoreid2} has proven to be more accurate and reliable. In \name, we adopt B-BOT+Attn-CL~\cite{pathak2019video} to prune coarse reID training data. Since it extracts an aggregated feature from four consecutive frames, we use a pre-defined aggregated feature distance threshold~\cite{pathak2019video} to determine if two \bboxes belong to the same person.

\subsection{Consistency Learning}
\label{subsec:noise-training}
DNN object detection models use backbone networks (\ie pre-trained classification networks like ResNet, GoogleNet) to extract discriminative feature maps from an image. The most commonly used method to retrain a DNN detector is to freeze backbone network and fine-tune the remaining detection layers (\ie RPN and RoI extractors) on a new dataset. In spite of fast convergence, this approach suffers from suboptimal performance due to the insufficiently trained backbone network. 

To address this limitation, we use \ncs training data to train backbone network. To be able to use pairs of \bboxes in \ncs training dataset, \name creates a reID-like pretext task. In specific, it takes pairs of images with bounding boxes belong to the same object as input and run them through the entire model to get feature maps. Afterwards, it calculates classification score (cls) of each image purely based on features within the paired \bbox, and use consistency loss in back propagation to train the backbone network (Figure~\ref{fig:new_design}). Consistency loss in \name is defined as $L_{consistency}=\sum_{k=1}^{P}CE(cls_{k}^{1}, cls_{k}^{2})$, where $P$ is the total number of pairs of \bboxes in two images, $CE$ represents cross entropy function, and $cls_{k}^{i}$ denotes predicted classification score of $k_{th}$ \bbox from $cam^i$. As consistency loss is minimized by fine-tuning, backbone network generates more representative feature maps.

After backbone network fine-tuning, we adopt the classic self-training framework~\cite{2019view1,2019selfOD1,2019RCNN} to update RPN and RoI feature extractor. Here \cs data (\ie confident pseudo-labels on the camera itself) is used directly as ground-truth for detection model fine-tuning. 

In summary, the overall loss function of \name\ can be formulated as $L_{overall}=\alpha*L_{consistency}+\beta*L^{det}$. We first minimize consistency loss through updating backbone network ($\alpha=1, \beta=0$) and reduce the loss of self-training by updating RPN and RoI feature extractor ($\alpha=0, \beta=1$). 

\section{Evaluation}
\label{sec:new_evaluation}
We evaluate \name on real-world multi-camera datasets and present evaluation highlights in this section. 


\subsection{Datasets}
\begin{table*}[!htb]
	\small
	\begin{center}
		\begin{tabular}{|c|c|c|c|c|c|}
			\hline
			\hline
			& Objects     & Total cameras & Resolution & Total frames & Avg. objects/frame \\ \hline
			WildTrack & Pedestrians & 7             & 1920*1080  & 29400        & 23                \\ \hline
			CityFlow  & Vehicles    & 5             & 960*480  & 9775         & 13                \\ \hline
			\hline
		\end{tabular}
	\end{center}
	\vspace{-10pt}
	\caption{Datasets description.}
	\vspace{-20pt}
	\label{table:dataset}
\end{table*}
Our experiments are conducted on two multi-camera detection datasets, namely WildTrack and CityFlow~\cite{wildtrack2018,aicity2018} \footnote{We used data collected from the first intersection of CityFlow.} (Table~\ref{table:dataset}). 
WildTrack is by far the largest multi-camera dataset for pedestrian detection and tracking while CityFlow is built for multi-camera vehicle tracking. 

\subsection{Settings}
We implemented \name\ with mmdetection~\cite{2019mmdetection}, an open source object detection toolbox based on Pytorch, and conducted all experiments using two Nvidia GeForce RTX 2080 Ti GPU.

\textbf{Models.} YOLOv3~\cite{yolov3} and Faster R-CNN~\cite{2015fasterrcnn} pre-trained on COCO with backbone of Darknet53 and ResNet101 are used as our default two-stage and three-stage object detection models. Evaluation results with different backbones are presented in Section~\ref{subsec:backbones}. $T_{cls}$ is set to $0.8$. We use SiamMask-E~\cite{chen2019fast} for tracking, and the state-of-the-art reID algorithms B-BOT-Attn-CL~\cite{pathak2019video} and VehicleNet~\cite{2020vehiclenet} for person and vehicle reID, respectively. 

\textbf{Training Settings.} We set the ratio of training set, evaluation set and testing set to $16:4:5$, 
batch size to $8$, and choose SGD with learning rate of $0.01$. All trainings last for $60$ epochs, with the first 30 epochs on backbone network and the subsequent 30 epochs on detection network.

\textbf{Evaluation Metrics.} We use mean Average Precision (mAP) over Intersection Over Union (IoU) of 0.8 (mAP@[0.8:1.0]).


\textbf{Baselines.} We compare \name with three self-supervised learning approaches. \\
i) \emph{Self-Training (ST)}: the most widely used self-training mechanism with confident pseudo-labels~\cite{2019RCNN,2019selfOD1}. It's trained in the supervised learning way with confident pseudo-labels. \\
ii) \emph{Self-Training with Gold Loss Correction (ST-GLC)}~\cite{hendrycks2018glc}: an improved version of ST with gold loss correction. It first estimates corruption matrix C of conditional corruption probabilities using confident pseudo-labels, and then uses C to correct class labels of all uncertain pseudo-labels. It uses all pseudo-labels to fine-tune the original object detection model. \\
iii) \emph{Self-Training with Consistency Loss (ST-CL)}~\cite{2019consistency1}: the most recent work on self-training. It uses two images (the original image and a flipped image) as input, and constructs consistency loss between two images during training. When we train on confident pseudo-labels, we use both supervised loss and consistency loss. When we train on uncertain pseudo-labels, we only use consistency loss. \\

In addition to the above three baselines, we report results from a supervised training with ground-truth (\ie to train custom detection model on each camera with human-labeled \bboxes from itself). It serves as an upper-bound of self-training methods. 
To show the gain of model customization, we also include mAP of the base model (\ie YOLOv3 and Faster R-CNN).

\subsection{End-to-End Results}
\begin{figure}[!htb]
	\begin{center}
		\begin{subfigure}{0.48\columnwidth}
			\includegraphics[width=0.9\columnwidth]{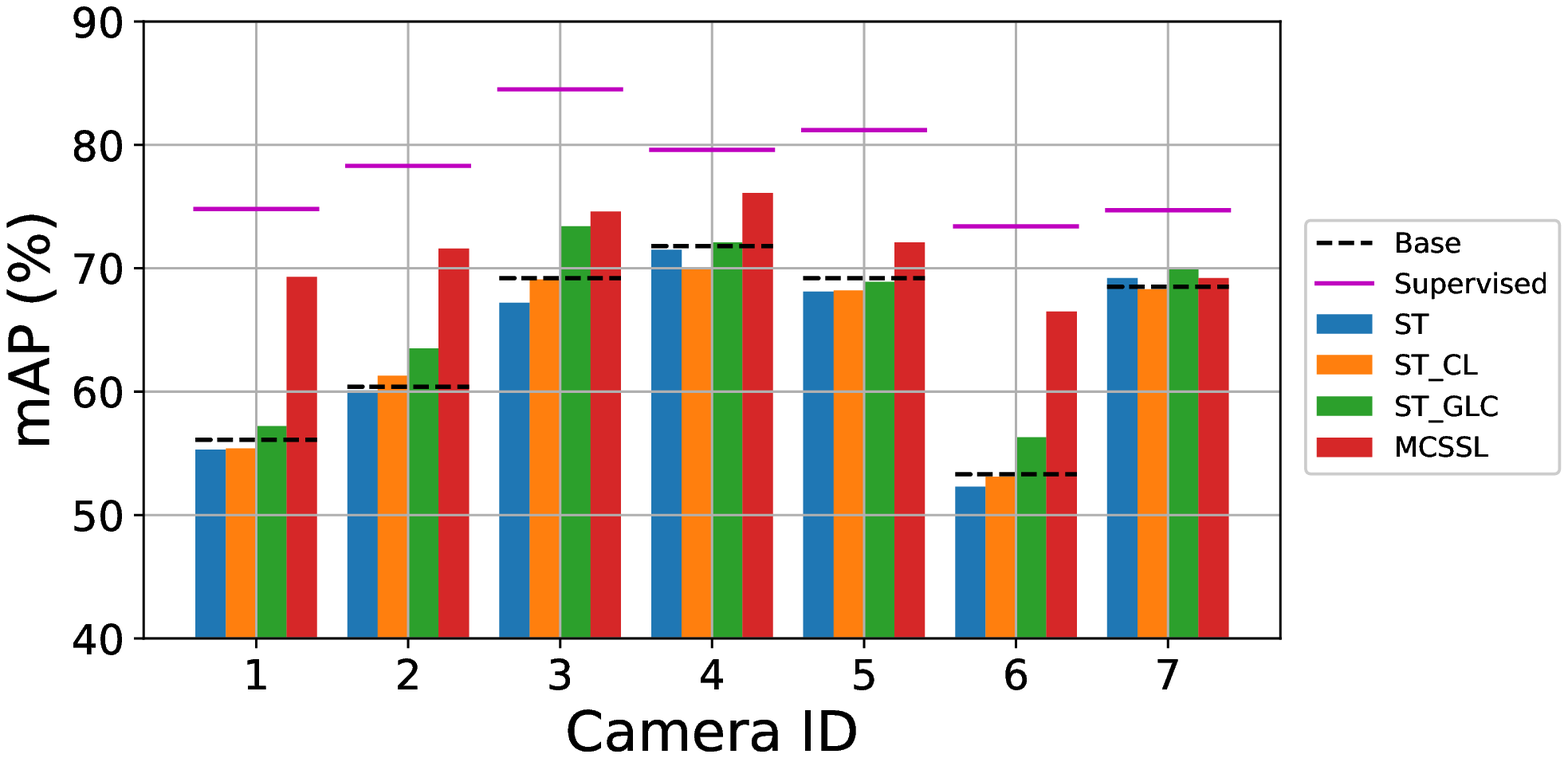}
			\caption{YOLOv3}
			\label{fig:sota_wildtrack_yolov3}
		\end{subfigure}
		\begin{subfigure}{0.48\columnwidth}
			\includegraphics[width=0.9\columnwidth]{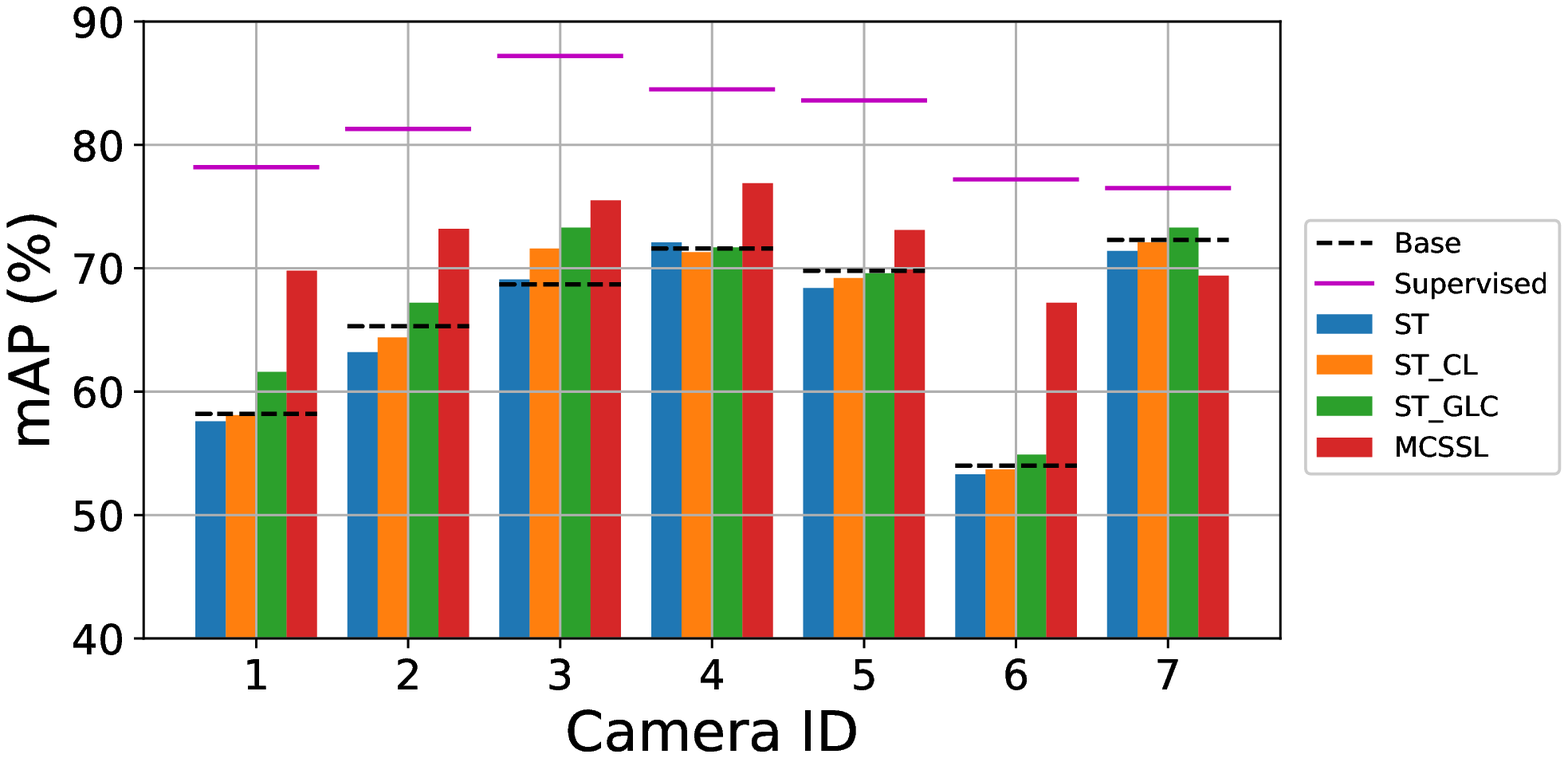}
			\caption{Faster R-CNN}
			\label{fig:sota_wildtrack_faster_rcnn}
		\end{subfigure}
	\end{center}
	\vspace{-10pt}
	\caption{mAP of self-training approaches on WildTrack.}
	\vspace{-10pt}
	\label{fig:sota_wildtrack}
\end{figure}

Figure~\ref{fig:sota_wildtrack} shows performances of customizing YOLOv3 and Faster R-CNN on WildTrack. Compared with the best known self-training approach (ST-GLC), \name\ improves the mAP of YOLOv3 and Faster R-CNN by $5.44\%$ and $4.78\%$ on average for each camera on WildTrack. This shows \name\ is an effective framework for both two-stage and three-stage object detection models. It's interesting to see that \name\ performs worse than ST-GLC on CAM-7 in both Figure~\ref{fig:sota_wildtrack_yolov3} and Figure~\ref{fig:sota_wildtrack_faster_rcnn}. This is due to the fact that CAM-7 has the least amount of shared FOV and hence much less \ncs training data (\eg $46.5\%$ less than other cameras on average in Figure~\ref{fig:sota_wildtrack_yolov3}) for backbone network fine-tuning. 

\begin{figure}[!htb]
	\begin{center}
		\begin{subfigure}{0.48\columnwidth}
			\includegraphics[width=0.9\columnwidth]{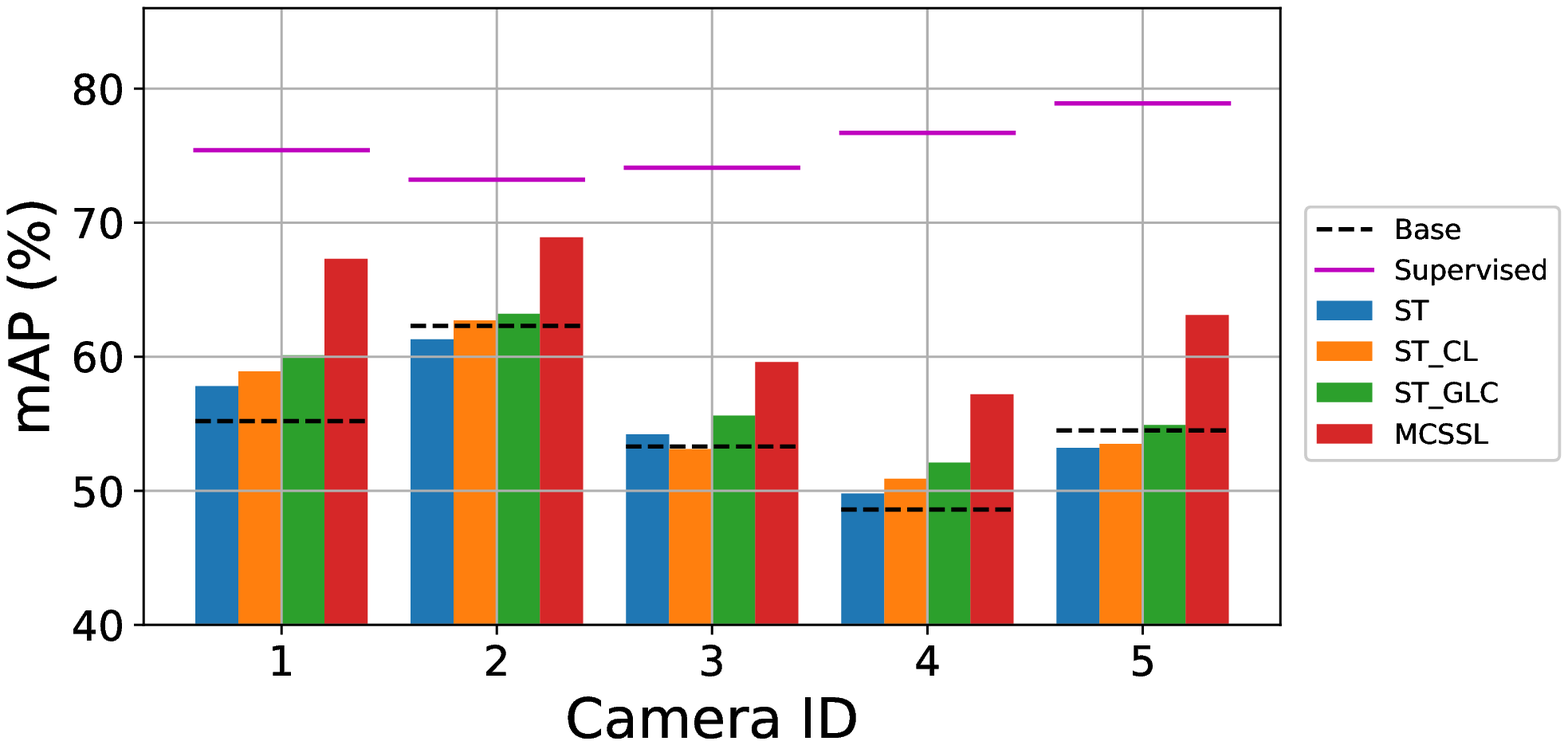}
			\caption{YOLOv3}
		\end{subfigure}
		\begin{subfigure}{0.48\columnwidth}
			\includegraphics[width=0.9\columnwidth]{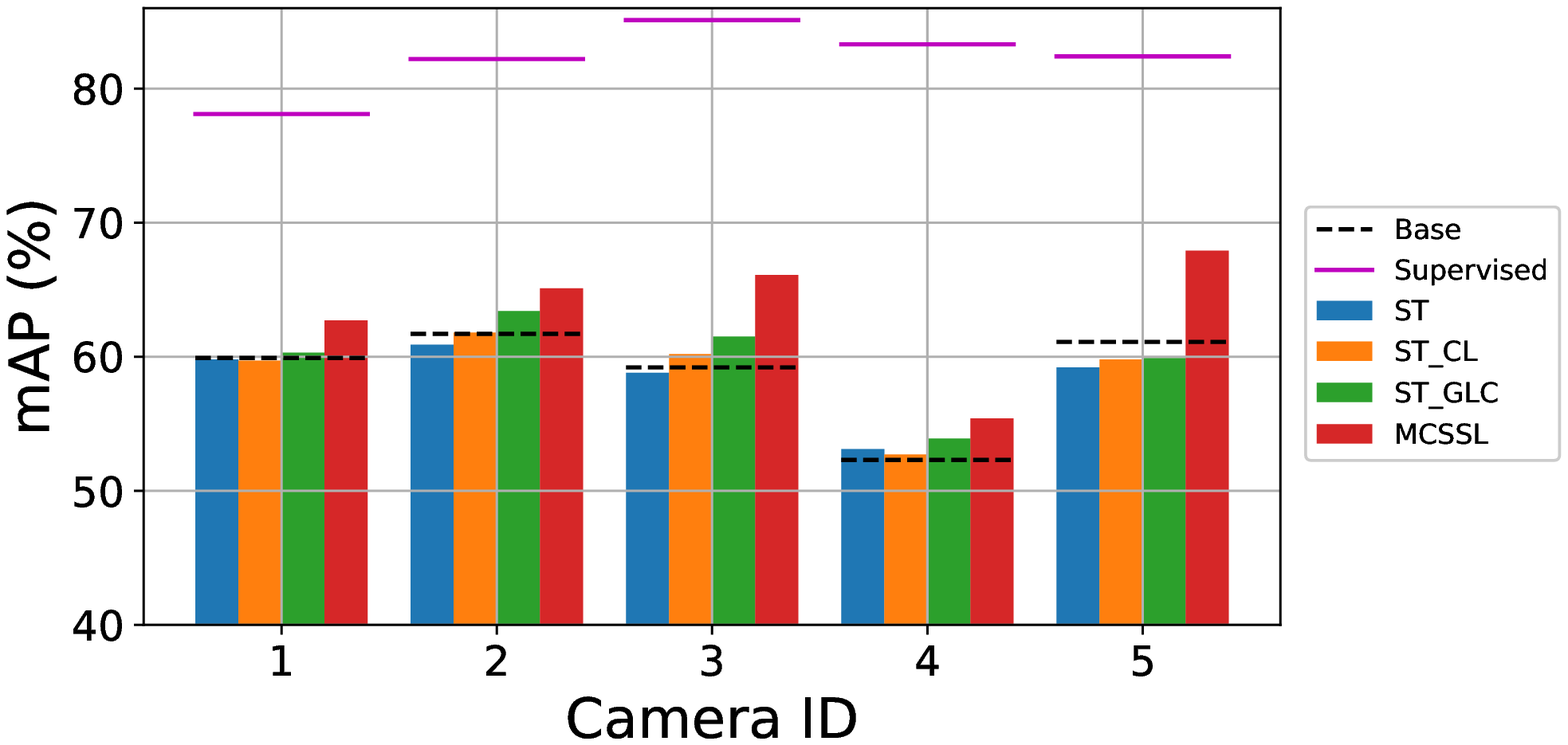}
			\caption{Faster R-CNN}
		\end{subfigure}
	\end{center}
	\vspace{-10pt}
	\caption{mAP of self-training approaches on CityFlow.}
	\vspace{-10pt}
	\label{fig:sota_cityflow}
\end{figure}

Using the same settings, we report performances of \name on YOLOv3 and Faster R-CNN on CityFlow in Figure~\ref{fig:sota_cityflow}. Compared with SL-GLC, \name obtains an average mAP improvement of $6.46\%$ and $3.64\%$ on two models. In particular, on CAM-5, YOLOv3 and Faster R-CNN are improved by $10.2\%$ mAP and $8.0\%$ mAP since CAM-5 largely overlaps with other cameras and hence gets more pseudo-labels from its neighbors.

\subsection{Ablation Study on Backbone Layers Fine-tuning}
\begin{figure}[!htb]
	\begin{center}
		\begin{subfigure}{0.48\columnwidth}
			\includegraphics[width=0.8\columnwidth]{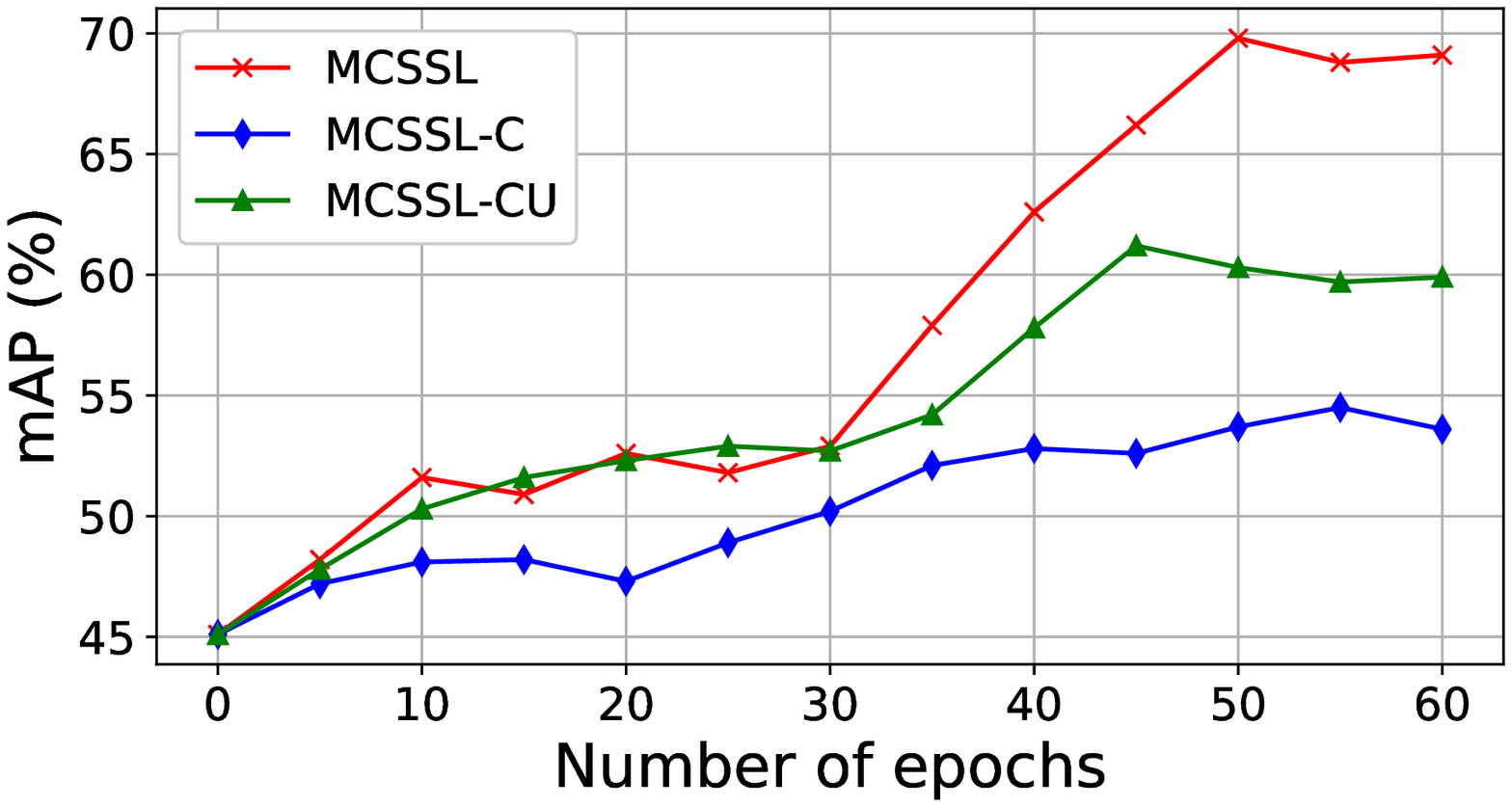}
			\caption{YOLOv3}
			\label{fig:yolov3_cam1}
		\end{subfigure}
		\begin{subfigure}{0.48\columnwidth}
			\includegraphics[width=0.8\columnwidth]{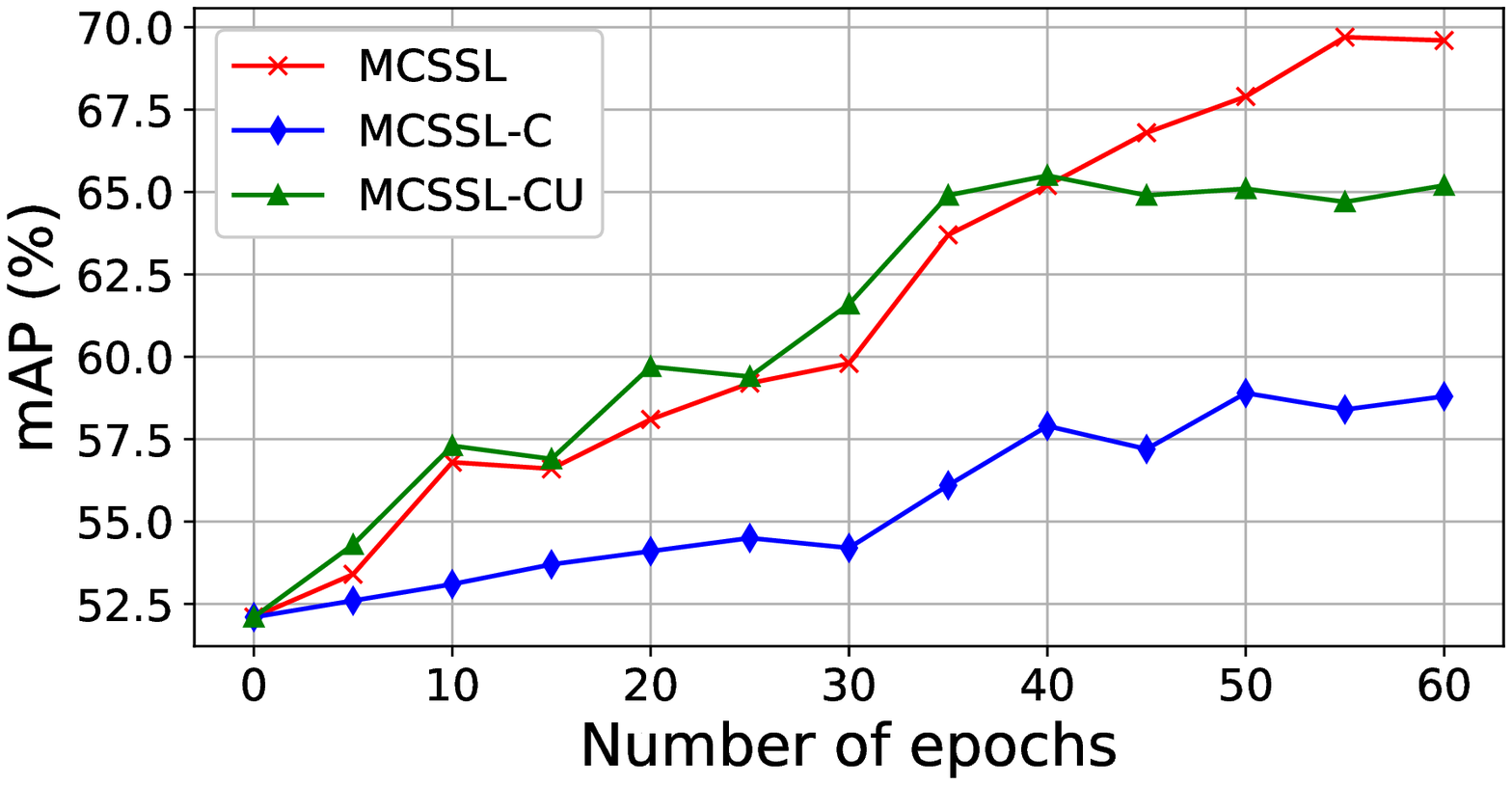}
			\caption{Faster R-CNN}
			\label{fig:faster_rcnn_cam1}
		\end{subfigure}
	\end{center}
	\vspace{-10pt}
	\caption{The performance comparison of different self-training processes over training epochs.}
	\vspace{-10pt}
	\label{fig:mole_cam1}
\end{figure}

To show how much uncertain \bboxes fit into the training of backbone layers, we compare three self-training processes on CAM-1 from WildTrack. \name\ only uses uncertain \bboxes to train backbone layers. \name-C trains backbone layers with confident \bboxes, and \name-CU uses both confident \bboxes and uncertain \bboxes.
We train backbone layers for 30 epochs and then fine-tune detection layers following the same protocol for another 30 epochs in all three approaches. In Figure~\ref{fig:mole_cam1}, as we expected, training backbone layers with more high-confident \bboxes (\name-CU) outperforms \name\ in the first 30 epochs. However, \name\ gets a better final mAP since training twice on confident \bboxes (in both backbone and detection layers fine-tuning) makes the detection model more prone to overfitting and hence limits its generalization ability on testing data. Compared with \name-C, we find that \name\ gets a better mAP in the first 30 epochs and a better final mAP in the later 30 epochs. This is because the size of confident \bboxes (\ie 1/4 of \name's training data) is insufficient to train the backbone network. Without powerful backbone layers, confident \bboxes cannot train detection layers effectively. In sum, the experiment validates our design of using uncertain \bboxes and confident \bboxes to train backbone layers and detection layers, respectively.
\label{subsec:ablation_study}
\subsection{Sensitivity Analysis}
\label{subsec:backbones}
\begin{figure}[!htb]
	\begin{center}
		\begin{subfigure}{0.48\columnwidth}
			\includegraphics[width=0.8\columnwidth]{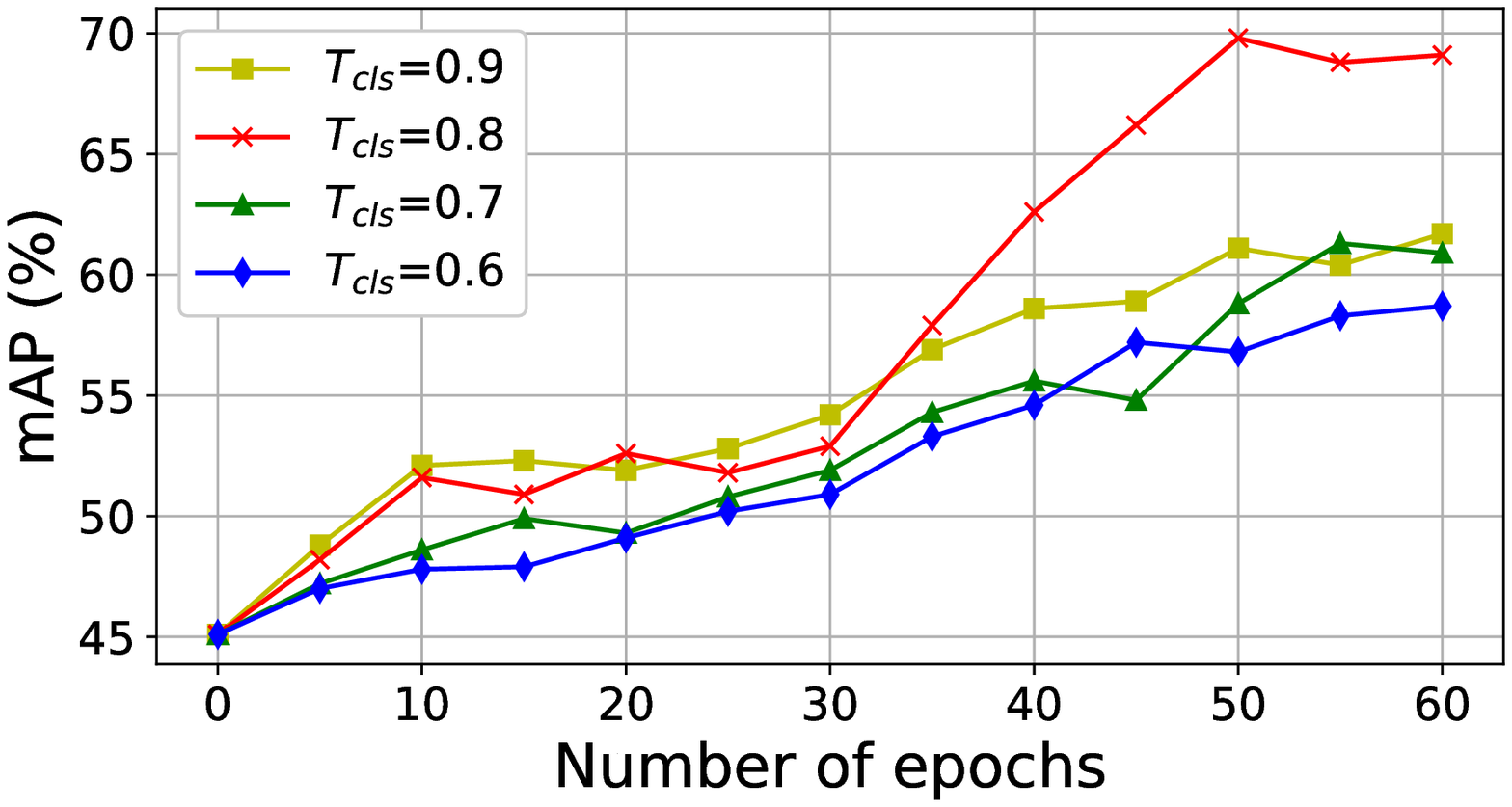}
			\caption{YOLOv3}
		\end{subfigure}
		\begin{subfigure}{0.48\columnwidth}
			\includegraphics[width=0.8\columnwidth]{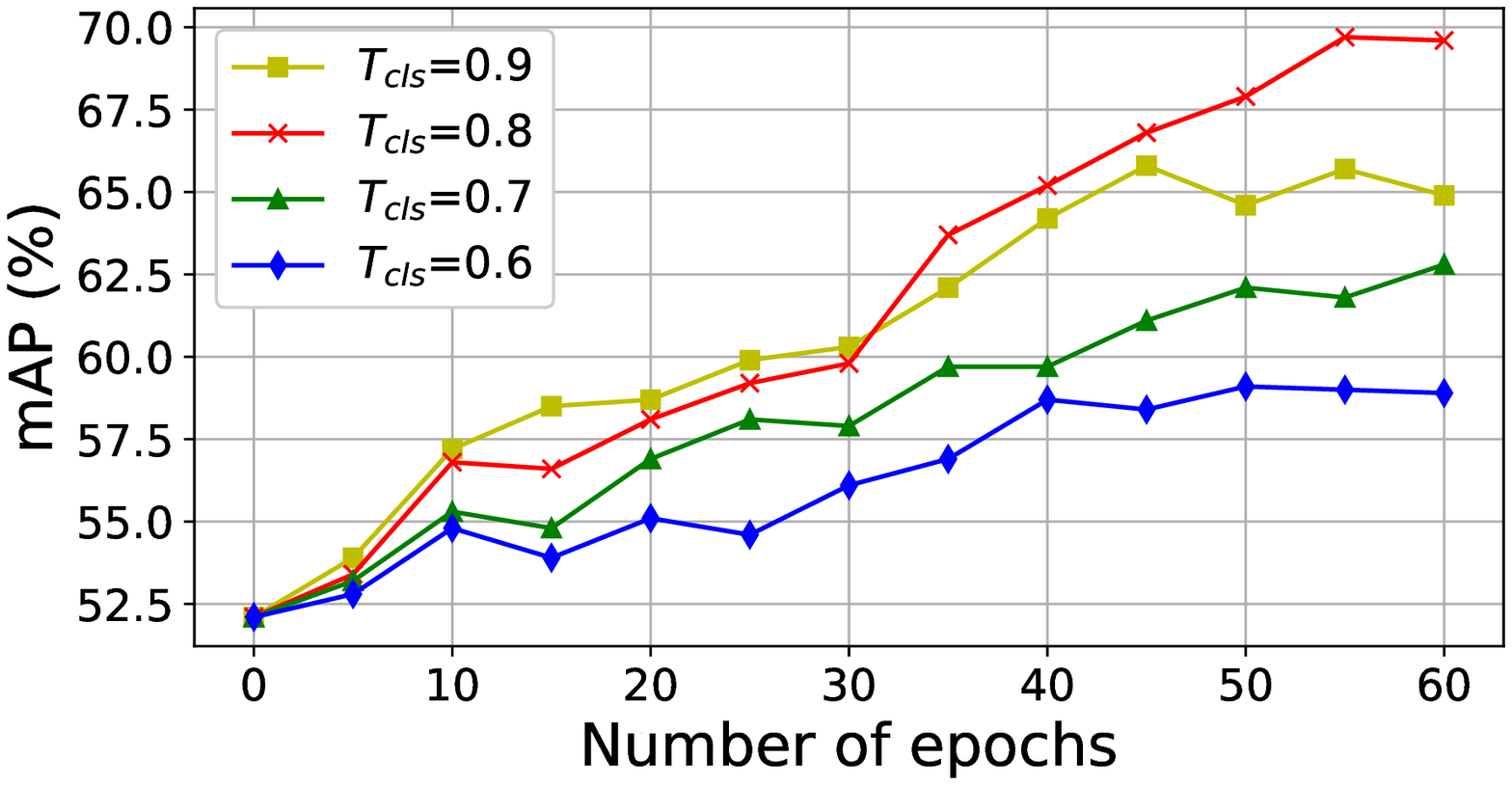}
			\caption{Faster R-CNN}
		\end{subfigure}
	\end{center}
	\vspace{-10pt}
	\caption{The performance comparison of \name\ under different $T_{cls}$ over training epochs.}
	\vspace{-10pt}
	\label{fig:mole_cam1_class}
\end{figure}

\begin{figure}[!htb]
	\begin{center}
		\begin{subfigure}{0.48\columnwidth}
			\includegraphics[width=0.8\columnwidth]{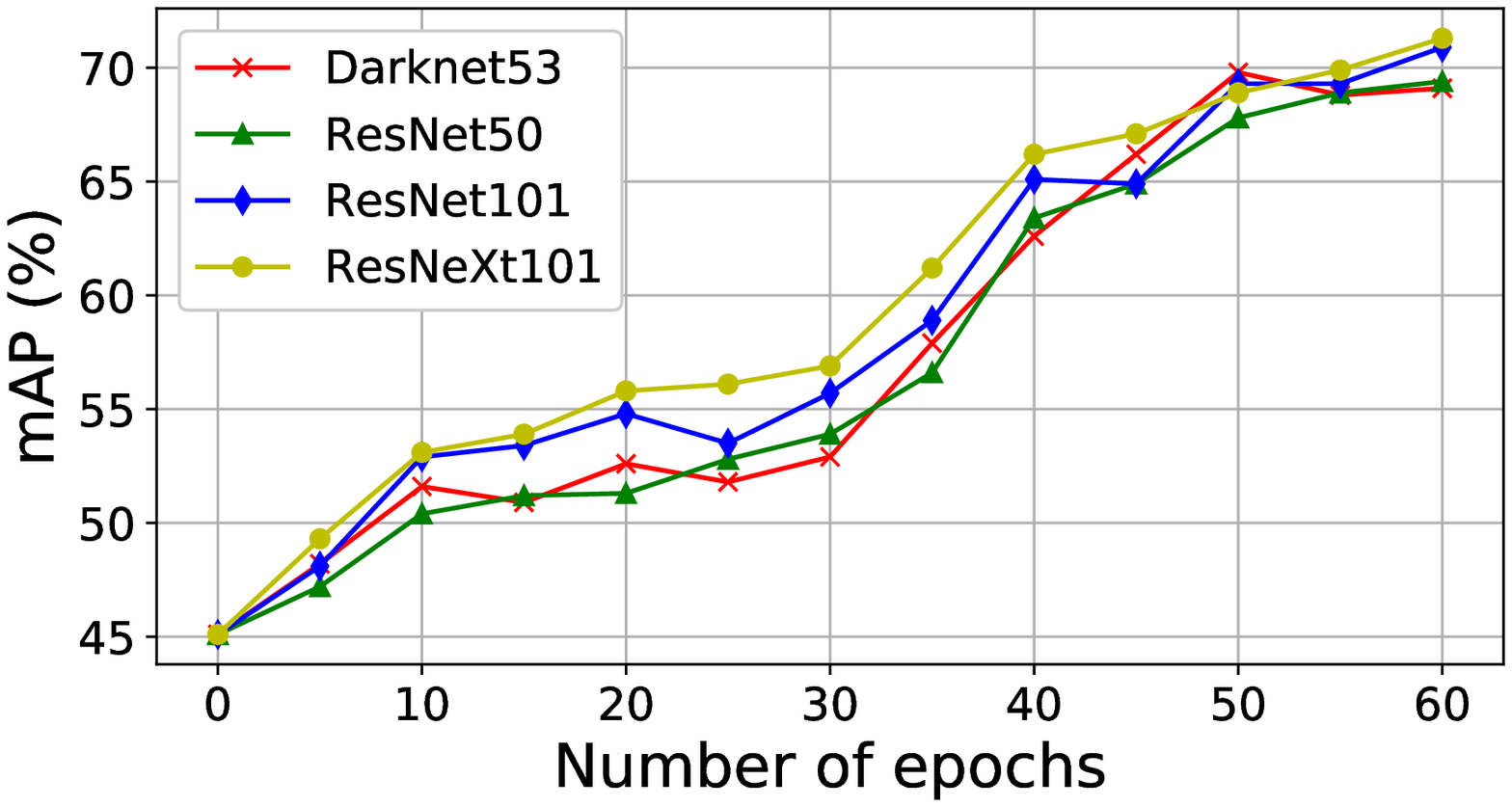}
			\caption{YOLOv3}
		\end{subfigure}
		\begin{subfigure}{0.48\columnwidth}
			\includegraphics[width=0.8\columnwidth]{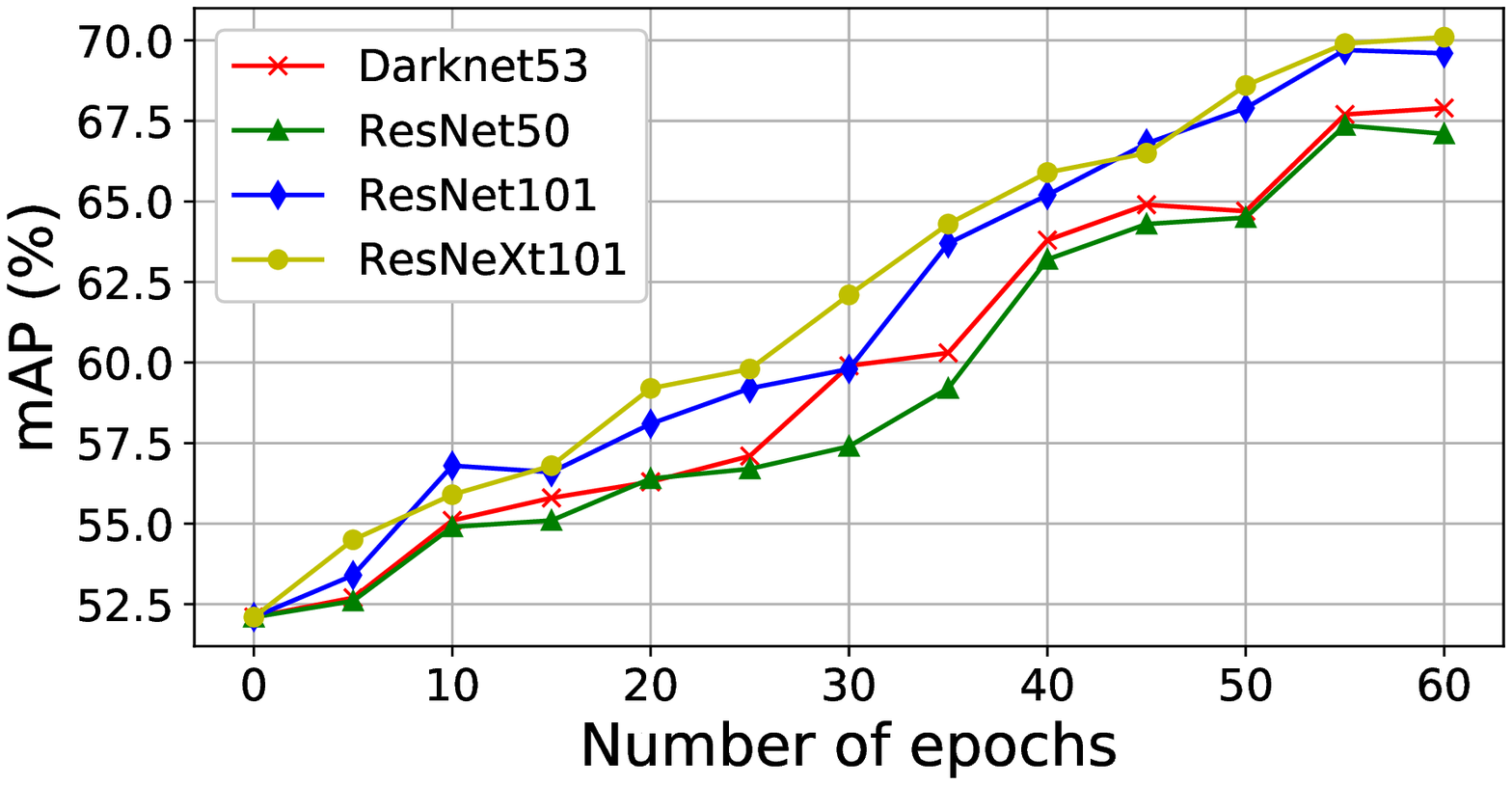}
			\caption{Faster R-CNN}
		\end{subfigure}
	\end{center}
	\vspace{-10pt}
	\caption{The performance comparison of \name\ with different backbones over training epochs.}
	\vspace{-10pt}
	\label{fig:mole_cam1_backbone}
\end{figure}

To analysis the impacts of important settings in \name, we conduct experiments on CAM-1 on WildTrack with various $T_{cls}$ (Figure~\ref{fig:mole_cam1_class}) and backbone architectures (Figure~\ref{fig:mole_cam1_backbone}). It comes as no surprise that \name yields lower accuracy with a smaller value of $T_{cls}$ due to the increasing noises in training data for detection network. However, it is worth noting that setting $T_{cls}$ too high (\eg $0.9$ in our experiment) could also negatively impact model customization due to the insufficient training of the detection network. We leave $T_{cls}$ as a hyperparameter to be tuned during training on different datasets. As shown in Figure~\ref{fig:mole_cam1_backbone}, \name is also amenable to different kinds of backbone networks as we see a steady improvement of detection accuracy with the increase of backbone capacity. 

\section{Conclusion}
%

We propose \name, a novel self-training mechanism with consistency loss, to customize object detection models in a multi-camera network. \name\ separates object detection models into backbone layers and detection layers and builds a reID-like pretext task for pre-train of backbone layers. To build training datasets, \name associates \bboxes between cameras using state-of-the-art tracking and reID algorithms with epipolar constraints. The large amount of \ncs data is used to train backbone network whereas high-quality \cs data is leveraged for detection network fine-tuning. Our evaluation results on two real-world datasets show \name can achieve the new state-of-the-art results for customizing detection models.
\\ \\

{\small
\bibliographystyle{named}
\bibliography{ijcai21}
}

\end{document}